\documentclass[10pt,twocolumn,letterpaper]{article}

\usepackage{wacv}
\usepackage{times}
\usepackage{epsfig}
\usepackage{graphicx}
\usepackage{amsmath}
\usepackage{amssymb}
\usepackage{mathrsfs}
\usepackage{ dsfont }
\usepackage{cleveref}
\usepackage{ subcaption}
\usepackage{algorithm}
\usepackage{algorithmic}
\usepackage[normalem]{ulem}
\useunder{\uline}{\ul}{}

\wacvfinalcopy % *** Uncomment this line for the final submission

\def\wacvPaperID{****} % *** Enter the wacv Paper ID here

\ifwacvfinal\fi
\begin{document}

%%%%%%%%% TITLE
\title{Targeted Attention for Generalized- and Zero-Shot Learning}

% Authors at the same institution
%\author{First Author \hspace{2cm} Second Author \\
% Authors at different institutions
\author{Abhijit Suprem\\
Georgia Institute of Technology\\
{\tt\small asuprem@gatech.edu}}

\maketitle
\ifwacvfinal\thispagestyle{empty}\fi

%%%%%%%%% ABSTRACT
\begin{abstract}
   The Zero-Shot Learning (ZSL) task attempts to learn concepts without any labeled data. Unlike traditional classification/detection tasks, the evaluation environment is provided unseen classes never encountered during training (in the related Generalized ZSL, or GZSL, setting, both seen and unseen classes are provided). As such, it remains both challenging and promising on a  variety of fronts, including unsupervised concept learning, domain adaptation, and dataset drift detection. Recently, there have been a variety of approaches towards solving ZSL, including improved metric learning methods, transfer learning, combinations of semantic and image domains using, e.g. word vectors, and generative models to model the latent space of known classes to classify unseen classes. We find many approaches require intensive training augmentation with attributes or features that may be commonly unavailable (attribute-based learning) or susceptible to adversarial attacks (generative learning). We propose a novel model with targeted attention with key modifications to ensure sufficiently improved performance in the ZSL setting without the need for feature or training dataset augmentation. We also incorporate approaches from the related person re-identification task (one-shot learning) for ZSL and GZSL.  We are able to achieve state-of-the-art performance on the CUB200 and Cars196 datasets in the ZSL setting compared to recent works, with NMI (normalized mutual inference) of 63.27 and top-1 of 61.04 for CUB200, and NMI 66.03 with top-1 82.75\% in Cars196. We also show state-of-the-art results in the GZSL setting, with Harmonic Mean R-1 of 66.14\% on the CUB200 dataset.

\end{abstract}

%%%%%%%%% BODY TEXT
\section{Introduction}
\label{sec:intro}
The traditional mode for image classification or object detection has relied on the key assumption that the training and prediction modes contain the same classes or targets. While this is sufficient for improving approaches for image detection, it is an impractical assumption for the real-world scenario where there may be limited labeled data or complete lack of labeled data. The former is the few-shot (or one-shot) recognition task, while the latter is the zero-shot learning (ZSL) task. Further, developing comprehensive models for all possible classes would require prohibitively large amounts of data, and training would be limited by the naturally long-tailed distribution of real-world object categories. 

There have been several approaches towards solving the ZSL task. We describe specifics of related work in \cref{sec:related}; here we give a brief overview to inform our proposed approach. Unsupervised metric learning has long been a staple of ZSL; more recently, \cite{dml} and \cite{proxy} develop variations of metric learning specifically for the ZSL task to improve class discrimination under adversarial training conditions as limited class instances. Since the common ZSL datasets CUB200 \cite{cub200} and Cars196 \cite{cars196} both provide class attributes and annotations, there have also been approaches that attempt transfer learning or co-learning of the image embeddings using attribute labels, with the assumption that real-world scenarios would also have such attributes freely available for unseen classes. More recently, generative models have been proposed in \cite{fgn} and \cite{izsl} to exploit capabilities of autoencoders and generative models in learning the latent space of a dataset. As such, the latent space representations are used to either augment training with synthetic examples or combined with class attributes for co-learning.

A common theme is augmenting the provided image data with additional, class-specific data to improve ZSL accuracy. However, such class-specific data may be unavailable under a real-world setting, or may themselves be new concepts never seen before. Further, the underlying distribution of both class attributes, and class labels may themselves face independent domain shift. Domain shift, also known as concept drift, is the phenomenon where the distribution of prediction data changes over time (a survey is available in \cite{driftb}, and an example of drift in text streams is explored in \cite{drifta}). 

In this paper, we propose an end-to-end approach for ZSL that does not rely on data augmentation. Our approach avoids reliance on class-attributes, which may not be common available. We also avoid training data augmentation with generative models since GANs without latent priors often exhibit \textit{holes} in the latent space caused by the irregular mapping from source to latent space, and GANs with latent priors (such as variational autoencoders) exhibit loss of information during the latent mapping. This loss of information makes them susceptible to the domain shift problem, since the latent model may have missed information that indicates change in concept or distribution. We describe these limitations in detail in \cref{sec:gans}.

To address these limitations, our approach is an end-to-end network for the ZSL and GZSL setting. Our model consists of a ResNet backbone with \textit{targeted} convolutional attention for discriminative fine-grained feature learning. We directly use the convolutional features for the clustering and classification instead of adding dense layers as in \cite{rkt} or \cite{dsp} in order to force both feature \textit{extraction} and feature \textit{interpretation} in the convolutional layers (normally, convolutional layers perform feature extraction and subsequent dense layers perform feature interpretation). Finally, we combine the Proxy NCA loss proposed in \cite{proxy} with a smoothed softmax functions during training only to improve discriminative ability. Our end-to-end ZSL model needs only the image to achieve state-of-the-art results on the CUB200 and Cars196 dataset, with NMI 63.27 and top-1 61.04 for CUB200 and NMI 66.03 with top-1 82.75\% on Cars196. We also test our model in the GZSL setting, where we achieve harmonic mean top-1 of 66.14\%.

\section{Related Work}
\label{sec:related}
In this section, we review recent literature on ZSL and GZSL and cover limitations of GANs in modeling the latent space.

\subsection{Zero-Shot Learning (ZSL)}
\label{sec:zslrelated}
Under the ZSL task, the training and prediction classes have no overlap; evaluation involves ability to cluster/classify unseen classes amongst themselves.

The lifted structured embedding approach (DML)  in \cite{dml} addresses the zero-shot learning problem with an improved metric learning loss function. Since metric learning suffers from the complementary problems of embedding collapse due to improper anchor-negative selection in mini-batch training and poor convergence in non-batched training, DML proposes  a structured loss function based on all positive and all negative pairs in the training set. %In each minibatch, the selected positive and negative pairs' embeddings and compared amongst each other and across pairs. Additionally, DML ensures hard negative selection within the mini-batch to avoid easy triplets.

Similarly, ProxyNCA \cite{proxy} also targets the loss function for addressing zero-shot learning. ProxyNCA proposes a set of learned proxies for improving metric learning. Under ProxyNCA, features from training-set images are mapped to the proxy-domain and the triplet loss (or any other distance loss function) is computed amongst the proxies themselves. %Intuitively, in each batch, images of each instances should map to the same proxy. As such, the loss over the proxies in a mini-batch is representative of the loss over the entire training set, leading to faster convergence.

iZSL, proposed in \cite{izsl} moves away from explicit metric learning to latent metric learning; in iZSL, a variational autoencoder (VAE) learns the latent space of seen classes and their attributes $a^*$. Assuming knowledge of these attributes in unseen classes, iZSL then, for each unseen instance, solves the following optimization problem: given instance $x$, transform $x$ to $z$ using the VAE encoder, then find the $a*$ in $z$ that maximize the VAE lower bound. This $a*$ are the class attributes that best represent the unseen instance $x$.

%. Since $z$ itself is represented as a gaussian mixture model of class attributes $\{a\}$, it suffices to ; in effect  Thus, any $x'$ with the same attributes as $x$ is of the same general class as $x$.

The Feature Generating Network \cite{fgn} (FGN) approach uses GANs similar to the iZSL approach; in contrast to iZSL, which imposes a variational prior for the latent space, FGN does not explicitly impose a prior on $z$ with the Kullback-Leibler divergence. Given unseen classes, synthetic features are used to generate a training set for classifying additional examples of the unseen classes.

%Synthetic images are generated from gaussian noise and an adversarial discriminator is used to improve generated features. 

The Discriminative Sampling Policy \cite{dsp} (DSP) approach targets the sampling policy of the triplet metric to improve hard negative selection while reducing instances of embedding collapse. The DSP approach develops a deep sampler that operates during model training to select effective training samples. Under DSP, an effective set of samples has high hardness, where hardness is a measure of the inter-class similarity (where objects of different classes have very similar features).

Finally, the Recurrent Knowledge Transfer \cite{rkt} (RKT) approach uses the provided class labels as part of a language model to transform the images to a semantic embedding space. RKT learns a distance metric for the image embedding space using the semantic embedding space, with the assumption that dissimilar words indicate dissimilar images and similar words indicate similar images. Word-to-word distance is computed using 500-dim word vectors.

\subsection{Generalized Zero-Shot Learning (GZSL)}
Under the GZSL task, both seen and unseen classes are provided during testing; evaluation measures r-1 on seen and unseen classes independently, as well as their harmonic mean.

Aligned variation autoencoders are proposed in \cite{cadavae} to co-learn both image features and the class embedding. The co-learning approach combines natural language descriptions of the provided image to cluster seen and unseen classes to their respective labels. The FGN approach described in the prior section also includes work on the GZSL setting, where generated features are used in conjunction with image features to discriminate seen and unseen classes. More recently, the Gradient Matching Generative Network (GMGN) in \cite{gmgn} generates synthetic images of unseen classes to augment the training data. Synthetic images are used to train an updated supervised model that includes both real samples from seen classes and synthetic embeddings samples from unseen classes. Differently from FGN, GMGN performs weak-supervision classification by using synthetic embeddings to drive unseen class detection.

\subsection{GANs}
\label{sec:gans}

Generative Adversarial Networks (GANs) were first introduced in \cite{gan} to produce synthetic images or text. The basic model for a GAN consists of two networks - the Generator and the Discriminator. The Generator $G(z)$ and Discriminator $D(x)$ are trained adversarially: the generator, given a random point $z\in\Re^z$, generates an image $x'$. The Discriminator, given a real image $x$ and a \textit{fake} image $x'$ attempts to distinguish between the two, completing the zero-sum min-max game:

\begin{multline}
\min_G \max_D V(D,G)=\mathds{E}_{\mathbf{x}\sim p_{data}(\mathbf{x})}[\log D(\mathbf{x})]    + \\
\mathds{E}_{\mathbf{z}\sim p_{z}(\mathbf{z})}[\log 1-D(G(\mathbf{z}))]
\end{multline}

A GAN can be combined with an Autoencoder (AE) to create an end-to-end latent modeler:
\begin{itemize}
	\item An Encoder $E$ that performs dimensionality reduction $E:\Re^n\rightarrow\Re^z, n >> z$, where $x\in\Re^n$
	\item A Generator $G$ that attempts to reconstruct $x'$ or a style-transfer image $y'\sim x'$ given $z'=E(x')$, therefore performing the reverse mapping $G:\Re^z\rightarrow\Re^n$
\end{itemize}

\begin{figure}[h]
	\centering
	\includegraphics[width=3.4in]{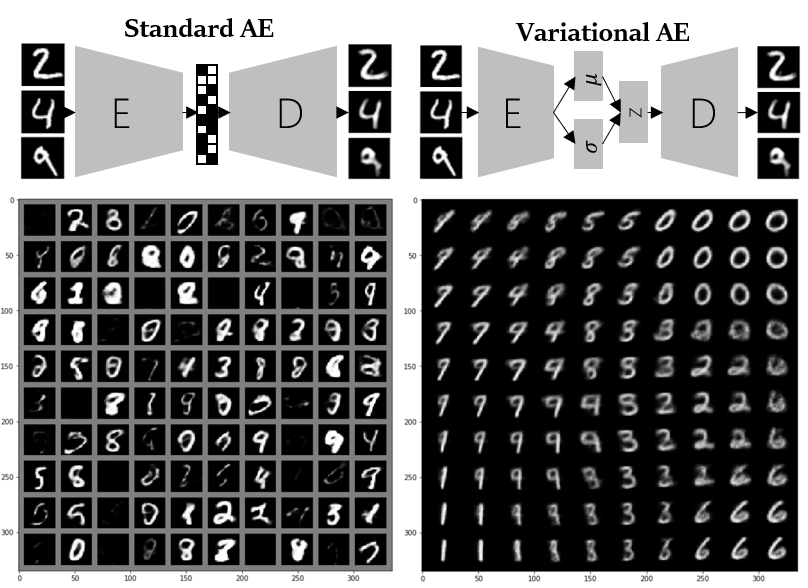}
	\caption{Autoencoders on MNIST. Both plots depict the latent space of autoencoders (AEs trained with 2D latent space). Each sub-image in the grid is from a linear distribution from $[0,0]$ to $[1,1]$. The latent representation on the left is from a SAE; without the smoothness constraint, there are no hybrid images indicating latent interpolation. In contrast, the VAE (right) has latent interpolation, indicating absence of holes in the latent space.}
	\label{fig:latentspace}
\end{figure}

However, the Encoder mapping $E:\Re^n\rightarrow\Re^z$ is irregular \cite{holes} due to the non-linearity of the encoding neural network, resulting in holes in the latent space that have not generalized. We show an example with MNIST in \cref{fig:latentspace}. In the standard autoencoder (left architecture in \cref{fig:latentspace}, with latent space of MNIST), generation of novel data (and conversely, encoding of novel data) is not possible due to the irregular mapping, resulting in holes in the latent spaces (black spots in generated grid).

Smoothness in the latent space can be enforced by adding a loss component to the training: matching the distribution of generated embeddings to a desired distribution using the Kullback Leibler divergence metric to create a Variational Autoencoder (VAE, shown in right architecture in \cref{fig:latentspace}). While VAEs can ensure the latent space is smooth, they lose significant discriminative ability in generation \cite{aae}, indicated by blurry synthetic images.

Our approach avoids synthetic feature generation for this reason, since the latent space is difficult to model and GAN training is unstable in drifting conditions.
%-------------------------------------------------------------------------

\section{Approach}
\label{sec:approach}
We develop an end-to-end network for the ZSL and GZSL setting using a single backbone network. Our approach implicitly learns relevant local and global features for unsupervised clustering without relying on data and feature augmentation or synthetic data. Data augmentation makes an assumption of available co-training and co-testing data that could be impractical in fully unconstrained zero-shot classification. Synthetic data generation creates, in the case of standard GAN/autoencoders, an irregular latent representation that would not capture novel information, and in the case of variational autoencoders, information loss that could limit novel information capture.

These existing approaches are designed to circumvent limitations in existing backbone networks (i.e. GoogLeNet, ResNet, VGG). Specifically, such networks are excellent at capturing in-distribution global and local features for discrimination, but fail in out-of-distribution tasks due to overfitting on training data. Co-testing with class attributes models mixture models on combinations of class attributes to enable novel class detection, while synthetic data generation attempts to increase training data for model updates. More specifically, the target is to allow the networks to model low-frequency/coarse features such as abstract shapes or colors while also learning on high-frequency detailed features for class-specific discrimination.

We achieve this implicitly without data augmentation or synthetic data generation with the following observations:

\begin{figure*}[t]
	\centering
	\includegraphics[width=6.8in]{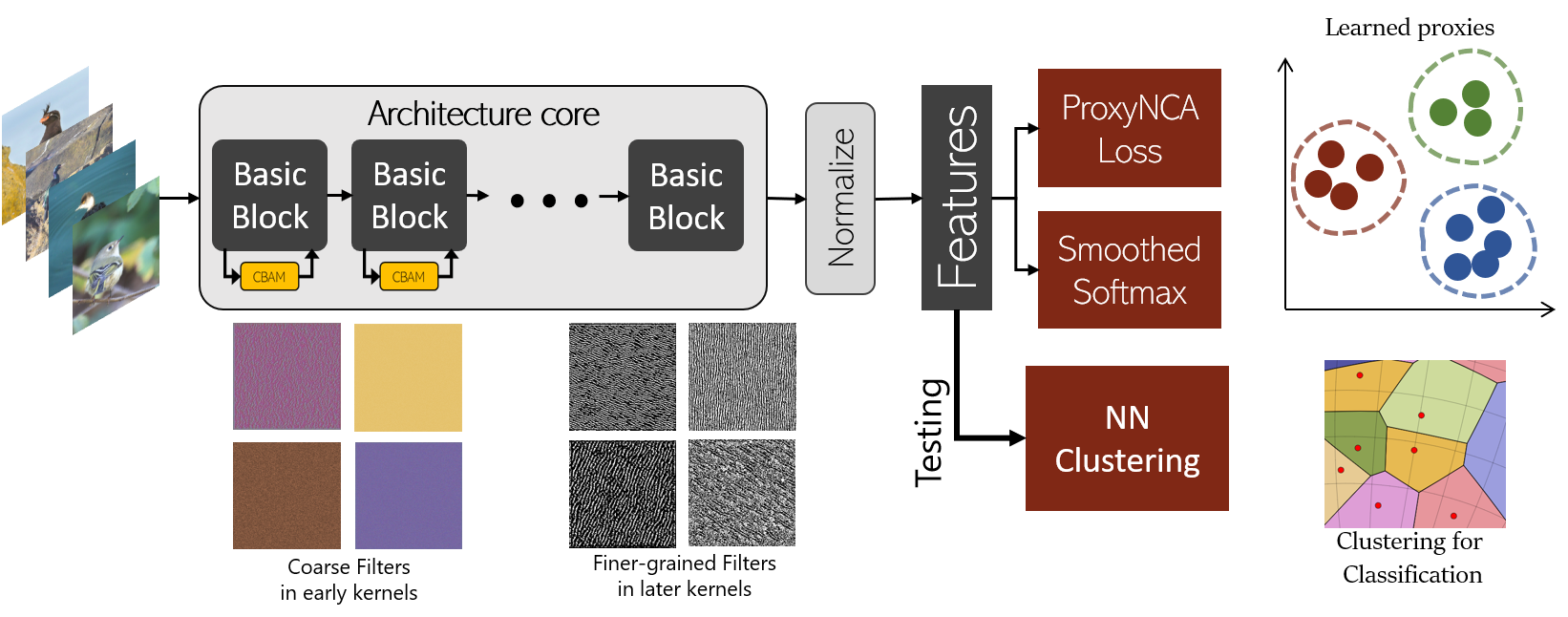}
	\caption{\textbf{Overall network architecture.} We use targeted attention within the architecture core (Resnets, or Inception) to improve coarse filters only. We use the ProxyNCA loss combined with a smoothed softmax loss to improve discriminative feature extraction. The softmax loss is used only during training. During testing, feature embeddings are clustered with nearest neighbor search. Best viewed in color.}
	\label{fig:architecture}
\end{figure*}

\begin{enumerate}
	\item It is well known that the earlier kernels in a convolutional network learn abstract, simple features such as colors. Some kernels also learn basic geometric shapes corresponding to image features in the low frequency range of images.
	\item Later kernels learn more class-specific details and extract detailed features corresponding to higher-frequency features. While these are useful for traditional image classification, they may overfit on the ZSL task.
	\item Convolutional layers are used for feature \textit{extraction}, and subsequent dense, or fully-connected layers, used for feature \textit{interpretation}. This forces the dense layers to learn image feature discrimination, instead of relying on convolutional filters. Since convolutional filters focus on nearby pixels with a spatial constraint, we believe relying on convolutional filters for feature interpretation to be more effective in tracking image invariant features.
\end{enumerate}

As such, current approaches have opted for backbones to perform discriminative, high-frequency learning, with data augmentation to reduce overfitting and allow for further generalization using e.g. co-learning or synthetic data generation. The adversarially-trained network in \cite{fgn} and \cite{dsp} uses hard examples to force the backbone network to learn high-frequency features on out-of-distribution samples. We have already noted limitations of these approaches.

We address these limitations in our approach by considering our earlier observations on convolutional filters: since the earlier filters learn low-frequency and coarse-grained features, we propose using the early kernels with attention modules. Since the later kernels learn high-frequency features, they do not require augmentation; otherwise they would begin to overfit on the training data. Thus, we use convolutional attention on the early kernels only, as opposed to attention throughout. We also address the loss of spatial image features in dense layers by removing them entirely and only use convolutional layers for both feature extraction and interpretation: given query and target, we evaluate their similarity on only convolutional features. In contrast to current approaches that use dense layers after the convolutional backbone to perform feature interpretation, we force the convolutional network to also learn feature interpretation simultaneously with feature extraction. 

\subsection{Network Overview}
\label{sec:overview}

In this section, we will describe our overall network and the targeted attention modules. We show our network in \cref{fig:architecture}. 

\begin{figure}[h]
	\centering
	\includegraphics[width=2.0in]{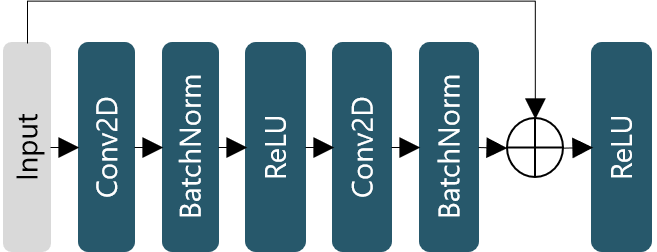}
	\caption{The ResNet basic block consists of 2 conv layers with normalization and relu, summed with the scaled and cropped residual from the input to the block.}
	\label{fig:basicblock}
\end{figure}

\paragraph*{Architecture Core} We test a variety of convolutional feature extractors such as ResNet, Inception, and VGG. We also examine variations of ResNet, including ResNet 18, ResNet34, and ResNet50. Each architecture core consists of several "basic blocks" chained together; a Resnet basic block is shown in \cref{fig:basicblock}. We apply targeted attention to these basic blocks, as opposed to each convolutional filter.

\begin{figure}[h]
	\centering
	\includegraphics[width=3.4in]{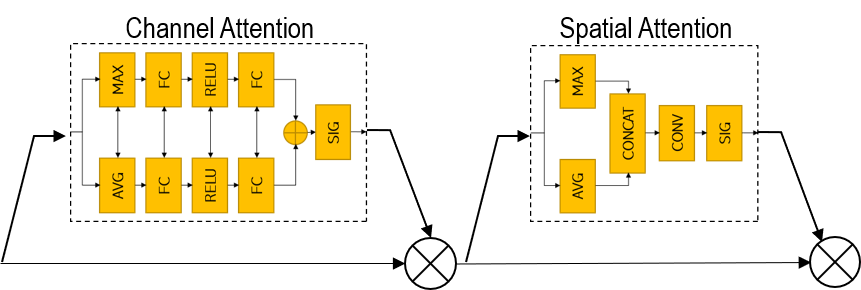}
	\caption{The convolutional block attention module consists of a spatial attention and a channel attention module to adjust feature maps during inference.}
	\label{fig:cbam}
\end{figure}

\paragraph*{Convolutional Attention} We use the CBAM block from \cite{cbam}, shown in \cref{fig:cbam} to add a attention module that can modify convolutional attention weights during inference. For the ResNet architectures. Since the earlier filters learn coarser features and the later filters learn fine-grained features, adding convolutional attention to all layers improves classification accuracy in general. However, in the one-shot or zero-shot tasks, CBAM on later layers causes networks to overfit on high-frequency features of the training set, reducing overall performance on the ZSL task (we examine this performance drop in \cref{sec:zslresults}). We propose adding attention only to the first basic block. Attention at the first basic blocks allows networks to learn discriminative coarse-grained filters for better class separation and metric learning, improving performance in the zero-shot setting.

\begin{figure}[h]
	\centering
	\includegraphics[width=2.0in]{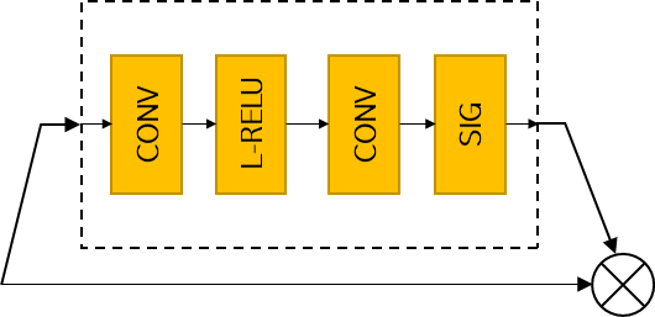}
	\caption{The Input Attention module regularizes the feature maps of the first convolutional filters outside the basic block in the architecture core.}
	\label{fig:iamodule}
\end{figure}

\paragraph*{Input Attention} Convolutional attention is not sufficient to improve generalization due to skewed feature maps in early convolutional filters. The first two convolutional filters are crucial in feature extraction since they occur at the beginning of the network path. We find that many feature maps at the first layers do not track any useful features; instead they either output random noise or focus on irrelevant features such as shadow. Recent work in \cite{dropout} has shown effectiveness of dropout as a regularization parameter for convolutional layers; however we noticed training instability with dropout in the first layer. Therefore, we develop the Input Attention module (IA), shown in \cref{fig:iamodule} to perform feature map regularization.

The IA module alleviates feature map skew by re-weighting feature weights. Unlike CBAM, which separates channel and spatial attention, we keep both in IA. The attention procedure uses two $3\times 3$ convolutional layers with Leaky ReLU activation to retain negative weights from the first convolutional layer in the architecture core. The output is passed through a sigmoid activation and element-wise multiplied with the input:

\begin{equation}
IA(F)=F\cdot Sig(f^{3\times 3} (LeakyReLU(f^{3\times 3} (F))))
\end{equation}

\begin{figure}[h]
	\centering
	\includegraphics[width=3.4in]{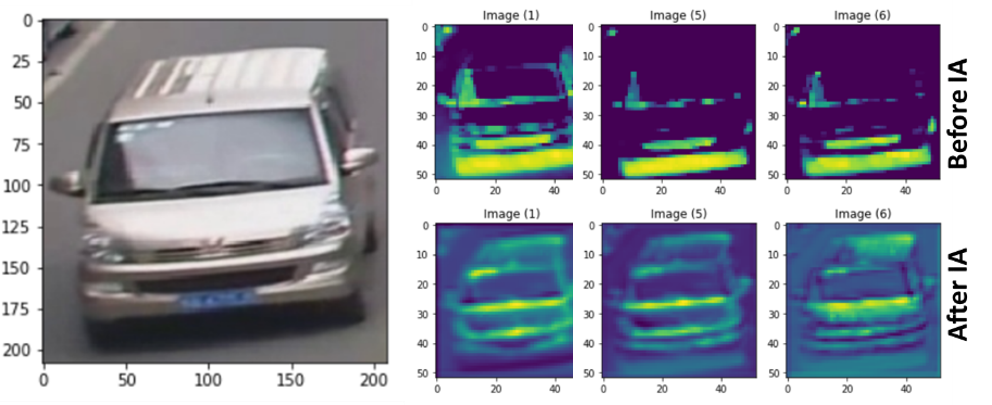}
	\caption{Feature map correction after Input Attention. Top features are skewed towards dark shadow, which does not carry any information. IA drives attention towards more useful details and corrects the skew.}
	\label{fig:ia}
\end{figure}

We avoid max and average pooling since they cause loss of information and we want to preserve discriminative features for the basic blocks in the architecture core. \cref{fig:ia} shows an example of feature map correction; the top layer shows original feature maps, which have skewed towards the shadow under the vehicle; the bottom layer shows the corrected feature map without skew.

\paragraph{Normalization}
We perform normalization on the extracted features from the architecture core for loss calculation. Since we are combining the ProxyNCA metric learning from \cite{proxy} and a softmax loss with label smoothing, we need to ensure the different loss targets do not conflict. Our use of the softmax loss in addition to a metric learning loss is inspired from the related person re-identification task, which is usually solved as a 1-shot learning task. In re-id, softmax loss is combined with a metric learning loss such as triplet loss to improve feature discrimination of the feature extractor.

In our network, since the proxy-based metric learning loss performs intra-class compaction and inter-class separation, it forces features into clusters in the latent space. The proxy loss is calculated as:

\begin{equation}
\label{eq:proxyloss}
\mathcal{L}_P=-\log (\frac{\exp(-d(x,p(y)))}{\sum_{p(z)\in p(Z)}\exp(-d(x,p(z)))})
\end{equation}

where $x$ is a data point functioning as the anchor from the traditional triplet loss and $p(y)$, $p(z)$ are the sampled positive and negative proxies for $x$. 

However, the label-smoothed softmax loss is formulated as

\begin{equation}
\label{eq:softsmooth}
\mathcal{L}_S=  \sum_i^N -q_i\log p_i
\end{equation}

We smooth the true labels such that:

\begin{equation}
\label{eq:smoothing}
q_i=\mathds{I}(\hat{y}=i)-(\epsilon\mathtt{sgn}(\mathds{I}(\hat{y}=i)-0.5))
\end{equation}

where we let the smoothing constant $\epsilon$ be proportional to number of classes in the training set $N$, so $\epsilon=1/N$. The smoothed softmax loss without a bias factor creates hyperplanes centered around the origin to separate classes. Since the loss targets are different (clusters for proxy NCA versus hyperplanes centered on the origin for softmax), we add a batch normalization layer between the features and the softmax loss calculation to separate the loss targets. Features before batch normalization are used for proxy loss calculation, and features after batch normalization are used for softmax loss calculation. During inference, the batch normalization layer is no longer necessary and features are used without normalization.

\subsection{Complete Network}
\begin{figure}[h]
	\centering
	\includegraphics[width=3.4in]{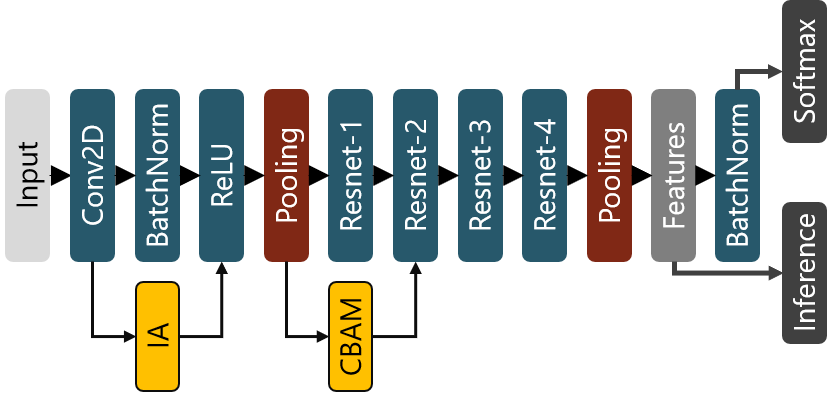}
	\caption{Complete network with ResNet backbone, targeted CBAM on coarse kernels, input attention, and feature normalization for softmax loss target separation.}
	\label{fig:fullresnet}
\end{figure}

A complete network incorporating targeted attention, input attention, and feature normalization with a ResNet backbone is shown in \cref{fig:fullresnet}. 

\section{Experimental Setup}
\label{sec:setup}
We will describe our experimental setup, including datasets, evaluation methods, and experiments performed.

\subsection{Datasets}
\label{sec:datasets}
We evaluate our approach on two common datasets: the Caltech-UCSD Birds-200-2011 CUB dataset \cite{cub200} and the Stanford Cars196 \cite{cars196}. We use the splits proposed in \cite{proxy}, where we choose 50\% of the classes as training classes and the remaining 50\% as testing classes. This is a harder problem than the 150-50 split for CUB200 since we use fewer classes to train the zero-shot learner ad evaluate on twice as many classes.

\paragraph*{CUB200} The CUB200 dataset contains 200 classes of birds. The CUB200 dataset is challenging for ZSL and GZSL since it contains fewer images per class than other datasets, with 60 samples per class on average. While class attributes such as names of the birds are provided, we do not make use of them in our end-to-end network, relying instead on only the images to perform zero-shot learning.

\paragraph*{Cars196} The Cars196 dataset contains 196 classes of vehicles. It is similarly challenging due to few images per class (on average, Cars196 has 82 images per class). Furthermore, vehicles exhibit a high degree of inter-class similarity as described in \cite{interclass} since most vehicles fit into a few form factors.

\subsection{Evaluation Metrics}
\label{sec:metrics}
We evaluate our models with two metrics: the normalized mutual information measure and the top-1 retrieval rate (we also show results for top-5 retrieval to compare to recent works).

\paragraph*{Normalized Mutual Information}
Normalized Mutual Information (NMI) measures clustering correlation between a predicted cluster set and ground truth cluster set; it is the ratio of mutual clustering information and the ground truth, and their harmonic mean. Given a set of predicted clusters $\Omega=\{c_1,c_2,\cdots ,c_k\}$ under a $k$-means clustering, we say that each $c_1$ contains instances determined to be of the same class. With ground truth clusters $\mathds{C}=\{c'_1, c'_2, \cdots, c'_m\}$, we calculate NMI as:

\begin{equation}
\label{eq:nmi}
NMI(\Omega, \mathds{C})=\frac{2I(\Omega,\mathds{C})}{H(\Omega)+H(\mathds{C})}
\end{equation}

Since NMI is invariant to label index, no alignment is necessary.

\paragraph*{Top-k}
We use the standard top-k ranking retrieval accuracy, calculated as the percentage of classes correctly retrieved at the first rank, and of those missed, percentage retrieved correctly on the second rank, and so on.

\subsection{Experiments}
\label{sec:experiments}
We examine performance impact of each of our proposed modules: the targeted attention module, input attention, and normalization, against a baseline feature extractor. We test in both the zero-shot and the generalized zero-shot setting for both CUB200 and Cars196. 

The \textbf{zero-shot learning} task performs evaluation on a testing set disjoint from the training set. As such, it is a measure of classification/clustering ability on unseen feature combinations. 

The \textbf{generalized zero-shot learning} (GZSL) task performs evaluation on testing set that contains both evaluation and training classes. It is inherently a harder problem since a feature extractor must be able to cluster unseen classes further away from seen classes, on which it may have overfit. It is also a more realistic scenario since real-world implementations often have a mixture of training and testing sets.

\section{Evaluation}
We evaluate performance of our end-to-end network on the CUB200 and Cars196 datasets. We build a baseline network with ResNet18 only without normalization, targeted attention, or input-attention. We use smoothed softmax loss combined with proxy loss in all cases, with total loss $\mathcal{L}=\lambda_P\mathcal{L}_P+\lambda_S\mathcal{L}_S$, where the loss weights are $\lambda_P=1$ and $\lambda_S$=0.5. We train our network with Adam optimizer with weight decay \cite{adamw} with learning rate $1e-4$ and decay $5e-4$. We use exponentially decaying learning rate with $\gamma=0.94$ and train for 50 epochs.

We will first cover results in the ZSL setting on each dataset. Then we will cover the GZSL setting results.

\subsection{ZSL Results}
\label{sec:zslresults}
\paragraph*{CUB200}

We evaluate NMI and R-1 on CUB200 and compare to recent approaches in \cref{tab:cubzsl}. Our baseline is similar to the Proxy-NCA method from \cite{proxy}; the addition of the softmax loss improves R-1 retrieval, though NMI suffers due to the increased overfitting from the softmax loss without feature normalization. With normalization, we see improvement to 62.5 NMI. Interestingly, addition of CBAM throughout the network actually reduces performance, since the later kernels begin overfitting on high-frequency features that appear exclusively on the training set. We test this intuition by applying targeted attention to just the first basic block (\texttt{B+CBAM\_E}), which we call \texttt{CBAM\_E} for CBAM-Early, and comparing to \texttt{CBAM\_L} or CBAM-Late, where CBAM is applied only to the final basic block. The results support our observations in \cref{sec:approach} - CBAM-Late has even worse performance, while CBAM-Early increases performance beyond CBAM everywhere. 

\begin{table}[h]
	\label{tab:cubzsl}
	\caption{Experimental results (\%) compared to recent approaches on the CUB-200-2011 dataset}
	\begin{tabular}{lrllll}
		\hline
		\multicolumn{1}{c}{\textbf{Method}}                                                                                          & \multicolumn{1}{c}{\textbf{NMI}}  & \textbf{R-1}  & \textbf{R-2}   & \textbf{R-4}  & \textbf{R-8} \\ \hline
		DML  \cite{dml}                                                                                                                        & \multicolumn{1}{c}{54.10}          & 42.80          & -              & -             & -            \\
		Proxy-NCA  \cite{proxy}                                                                                                                      & 59.53                             & 49.21         & 61.90           & 67.90          & 72.40         \\
		DSP  \cite{dsp}                                                                                                                        & 61.70                              & 53.60          & 65.50           & 76.90          & -            \\
		iZSL-R \cite{izsl}                                                                                                                        & -                                 & 57.42         & -              & -             & -            \\
		iZSL  \cite{izsl}                                                                                                            & -                                 & 55.37         & -              & -             & -            \\
		FGN \cite{fgn}                                                                                                                        & -                                 & 57.30          & -              & -             & -            \\ \hline
		Baseline (B)                                                                                                                 & 58.51                            & 54.58          & 66.89          & 77.54          & 85.72         \\
		B+Norm                                                                                                                       & 62.55                              & 58.59          & 70.38          & 80.05          & 87.65         \\
		B+CBAM                                                                                                                       & \multicolumn{1}{l}{56.69}         & 52.70          & 64.76          & 75.71         & 84.18        \\
		B+CBAM\_L                                                                                                                    & \multicolumn{1}{l}{40.08}         & 20.28         & 29.70         & 41.01         & 52.96        \\
		B+CBAM\_E                                                                                                                    & \multicolumn{1}{l}{63.27}          & 61.04           & \textbf{72.46} & \textbf{82.18} & 88.93        \\ \hline
		\textbf{\begin{tabular}[c]{@{}l@{}}Baseline\\ \,+CBAM\_E\\ \,+IA\\ \,+Norm\end{tabular}} & \multicolumn{1}{l}{\textbf{64.30}} & \textbf{61.33} & 72.32          & 81.72        & \textbf{89.99}  \\ \hline
	\end{tabular}
\end{table}

\paragraph*{Cars196}
We also evaluate NMI and R-1 on Stanford Cars196 dataset and compare to recent approaches in the following table. We note similar trend to CUB200, where we find CBAM-Late has a detrimental effect on the overall network's ability to differentiate novel classes. We find this as further evidence of the impact later kernels' high frequency feature extraction have on the ZSL task. On our baseline combined with CBAM-Early, Input Attention, and batch normalization, we achieve state-of-the-art results, with NMI 66.03\% and Rank-1 retrieval of 82.75\%, an improvement of over 10\%.

\begin{table}[h]
	\label{tab:carszsl}
	\caption{Experimental results (\%) compared to recent approaches on the Cars196 dataset}
	\begin{tabular}{llllll}
		\hline
		\multicolumn{1}{c}{\textbf{Method}}                                          & \multicolumn{1}{c}{\textbf{NMI}} & \textbf{R-1}   & \textbf{R-2}   & \textbf{R-4}   & \textbf{R-8}   \\ \hline
		DML \cite{dml}                                                                         & \multicolumn{1}{c}{56.70}        & 49.50          & -             & -             & -             \\
		DSP  \cite{dsp}                                                                        & \multicolumn{1}{r}{64.40}        & 72.90          & 81.60          & 88.80          & -             \\
		Proxy-NCA \cite{proxy}                                                                       & \multicolumn{1}{r}{64.90}        & 73.22          & 82.42          & 86.36          & 88.68          \\ \hline
		Baseline (B)                                                                 & \multicolumn{1}{r}{54.63}        & 72.28          & 81.63          & 88.52          & 93.27          \\
		B+Norm                                                                       & \multicolumn{1}{r}{64.59}        & 80.40          & 87.49          & 92.41          & 95.54          \\
		B+CBAM                                                                       & 56.71                            & 73.55          & 82.09          & 87.99          & 92.26          \\
		B+CBAM\_L                                                                    & 31.39                            & 22.94          & 32.72          & 43.56          & 56.08          \\
		B+CBAM\_E                                                                    & 63.04                            & 80.56          & 87.86          & 92.52          & 95.73          \\ \hline
		\textbf{\begin{tabular}[c]{@{}l@{}}Baseline\\ \,+CBAM\_E\\ \,+IA\\ \,+Norm\end{tabular}} & \textbf{66.03}                   & \textbf{82.75} & \textbf{89.68} & \textbf{93.72} & \textbf{96.41} \\ \hline
	\end{tabular}
\end{table}

\paragraph*{Discussion}

We made three key observations about conv-net backbones:
(i) since early kernels learn abstract features, they are important in generalizing, (ii) since later kernels learn class-specific features, they are prone to overfitting on their training set's feature distribution, and (iii) reliance on dense layers for metric learning causes feature \textit{interpretation}, where the relative importance of features is learned, to occur in dense layers, which do not capture an image's spatial characteristics the way convolutional kernels do.

Under these observations, we proposed four explicit modifications to address these pitfalls: targeted attention to improve early kernels, input-attention too regularize early kernels and increase the number of usable kernels, a softmax loss to improve feature discrimination, and a normalization layer to separate softmax loss targets and the metric learning loss targets. We also propose an implicit modification - using no dense layers in our feature extractor.

Our results clearly indicate the impact attention has at different layers in the feature extractor. Attention only at high-frequency feature extractors (i.e. later kernels) only increases in-distribution performance, to the detriment of out-of-distribution generalization. Consequently, attention throughout also has poor performance since late attention reduces the efficacy on ZSL. However, a single attention module on the first basic block, as described in \cref{fig:fullresnet}, delivers significant performance improvement compared to the baseline in both experiments. The lack of dense layers also aids us in performance, since the baseline model (a fully convolutional feature extractor with metric learning loss and softmax loss) is competitive with state-of-the-art methods that use intensive data augmentation or co-learning.

\subsection{GZSL Results}
In the GZSL setting, both seen and unseen classes are provided during the evaluation phase. We evaluate at test time the accuracy on seen classes $\mathbf{s}$ and unseen classes $\mathbf{u}$, along with the harmonic mean $\mathbf{H}=(2\cdot u\cdot s)/(u+s)$. We perform evaluation on CUB200 only, as there is limited work on GZSL in Cars196.

\begin{table}[h]
	\label{tab:gzsl}
	\caption{Experimental results (\%) in the GZSL setting on the CUB200 dataset}
	\centering
	\begin{tabular}{lrll}
		\hline
		\multicolumn{1}{c}{\textbf{Method}} & \multicolumn{1}{c}{\textbf{u}}     & \multicolumn{1}{c}{\textbf{s}}     & \multicolumn{1}{c}{\textbf{H}}     \\ \hline
		ReViSE \cite{revise}                             & \multicolumn{1}{c}{28.30}          & 37.60          & 32.30          \\
		Adv-GZSL \cite{agzsl}                            & 31.50                              & 40.20          & 35.30          \\
		VisRep \cite{visrep}                             & 28.80                              & 55.70          & 38.00          \\
		Cycle-WGAN \cite{cyclewgan}                          & 46.00                              & 60.30          & 52.20          \\
		CADA-VAE \cite{cadavae}                           & 53.50                              & 51.60          & 52.40          \\
		FGN  \cite{fgn}                               & \multicolumn{1}{l}{50.30}          & 58.30          & 54.00          \\
		GMGN \cite{gmgn}                               & \multicolumn{1}{l}{\textbf{57.90}} & 71.20          & 63.90          \\ \hline
		\textbf{Ours}                       & \multicolumn{1}{l}{{\ul 54.44}}    & \textbf{84.26} & \textbf{66.14} \\ \hline
	\end{tabular}
\end{table}

\paragraph*{Discussion}
We achieve state-of-the-art results in the GZSL setting with our end-to-end network with CBAM-Late, input attention, and batch normalization. As expected, the performance on seen classes is higher than performance on unseen classes. Our results indicate the performance improvements we make in the GZSL setting compared to existing approaches with our targeted attention module. In contrast, existing approaches require data augmentation, co-learning, or adversarial learning. Our results are competitive using our proposed targeted convolutional attention, normalization, and input-attention.

\section{Conclusions}
\label{sec:conc}

In this work, we have presented an end-to-end model for the zero-shot and generalized zero-shot learning tasks. Our approach achieves state-of-the-art results in both tasks. Furthermore, our approach is novel from existing approaches as it does not use data augmentation or co-learning from attributed features; instead we achieve competitive results with only the provided images on our evaluation datasets.

We developed our model with insights from the related one-shot problem of person and vehicle re-identification, and by making several key observations about the nature of convolutional networks and their roles as feature extractors. Specifically, we adopt the practice of combining a metric learning loss with a softmax loss during the training procedure. The metric learning loss teaches the network to differentiate between classes, while the softmax loss trains the network to extract more discriminative features. 

In conjunction, we make some intuitive changes to the standard backbone: we remove all dense layers to force the convolutional filters to perform both feature extraction and feature interpretation. We also examined the effects of targeted attention: with the observation that earlier kernels learn coarse features and later kernels learn fine-grained, \textit{high-frequency} features, we conclude the ZSL and GZSL tasks are best served by models that avoid overfitting on the training set. While recent approaches attempt this with adversarial training or data augmentation, we propose using targeted attention at the earlier filters to force the network to focus on coarse features, as opposed to fine-grained features. 

Our proposed model incorporating these changes achieves state-of-the-art results on both the CUB200 and Cars196 dataset in the ZSL setting. We also show that our approach achieves state-of-the-art performance on CUB200 dataset in the GZSL setting as well.

\pagebreak

{\small
\bibliographystyle{ieee}
\bibliography{main}

\begin{thebibliography}{10}\itemsep=-1pt

\bibitem{visrep}
M.~Bucher, S.~Herbin, and F.~Jurie.
\newblock Generating visual representations for zero-shot classification.
\newblock In {\em Proceedings of the IEEE International Conference on Computer
  Vision}, pages 2666--2673, 2017.

\bibitem{dsp}
Y.~Duan, L.~Chen, J.~Lu, and J.~Zhou.
\newblock Deep embedding learning with discriminative sampling policy.
\newblock In {\em Proceedings of the IEEE Conference on Computer Vision and
  Pattern Recognition}, pages 4964--4973, 2019.

\bibitem{cyclewgan}
R.~Felix, V.~B. Kumar, I.~Reid, and G.~Carneiro.
\newblock Multi-modal cycle-consistent generalized zero-shot learning.
\newblock In {\em Proceedings of the European Conference on Computer Vision
  (ECCV)}, pages 21--37, 2018.

\bibitem{driftb}
J.~Gama, I.~{\v{Z}}liobait{\.e}, A.~Bifet, M.~Pechenizkiy, and A.~Bouchachia.
\newblock A survey on concept drift adaptation.
\newblock {\em ACM computing surveys (CSUR)}, 46(4):44, 2014.

\bibitem{gan}
I.~Goodfellow, J.~Pouget-Abadie, M.~Mirza, B.~Xu, D.~Warde-Farley, S.~Ozair,
  A.~Courville, and Y.~Bengio.
\newblock Generative adversarial nets.
\newblock In {\em Advances in neural information processing systems}, pages
  2672--2680, 2014.

\bibitem{interclass}
T.-W. Huang, J.~Cai, H.~Yang, H.-M. Hsu, and J.-N. Hwang.
\newblock Multi-view vehicle re-identification using temporal attention model
  and metadata re-ranking.
\newblock In {\em AI City Challenge Workshop, IEEE/CVF Computer Vision and
  Pattern Recognition (CVPR) Conference, Long Beach, California}, 2019.

\bibitem{cars196}
J.~Krause, M.~Stark, J.~Deng, and L.~Fei-Fei.
\newblock 3d object representations for fine-grained categorization.
\newblock In {\em Proceedings of the IEEE International Conference on Computer
  Vision Workshops}, pages 554--561, 2013.

\bibitem{adamw}
I.~Loshchilov and F.~Hutter.
\newblock Fixing weight decay regularization in adam.
\newblock {\em arXiv preprint arXiv:1711.05101}, 2017.

\bibitem{aae}
A.~Makhzani, J.~Shlens, N.~Jaitly, I.~Goodfellow, and B.~Frey.
\newblock Adversarial autoencoders.
\newblock {\em arXiv preprint arXiv:1511.05644}, 2015.

\bibitem{proxy}
Y.~Movshovitz-Attias, A.~Toshev, T.~K. Leung, S.~Ioffe, and S.~Singh.
\newblock No fuss distance metric learning using proxies.
\newblock In {\em Proceedings of the IEEE International Conference on Computer
  Vision}, pages 360--368, 2017.

\bibitem{dml}
H.~Oh~Song, Y.~Xiang, S.~Jegelka, and S.~Savarese.
\newblock Deep metric learning via lifted structured feature embedding.
\newblock In {\em Proceedings of the IEEE Conference on Computer Vision and
  Pattern Recognition}, pages 4004--4012, 2016.

\bibitem{dropout}
S.~Park and N.~Kwak.
\newblock Analysis on the dropout effect in convolutional neural networks.
\newblock In {\em Asian Conference on Computer Vision}, pages 189--204.
  Springer, 2016.

\bibitem{gmgn}
M.~B. Sariyildiz and R.~G. Cinbis.
\newblock Gradient matching generative networks for zero-shot learning.
\newblock In {\em Proceedings of the IEEE Conference on Computer Vision and
  Pattern Recognition}, pages 2168--2178, 2019.

\bibitem{cadavae}
E.~Schonfeld, S.~Ebrahimi, S.~Sinha, T.~Darrell, and Z.~Akata.
\newblock Generalized zero-and few-shot learning via aligned variational
  autoencoders.
\newblock In {\em Proceedings of the IEEE Conference on Computer Vision and
  Pattern Recognition}, pages 8247--8255, 2019.

\bibitem{drifta}
A.~Suprem, A.~Musaev, and C.~Pu.
\newblock Concept drift adaptive physical event detection for social media
  streams.
\newblock In {\em World Congress on Services}, pages 92--105. Springer, 2019.

\bibitem{revise}
Y.-H.~H. Tsai, L.-K. Huang, and R.~Salakhutdinov.
\newblock Learning robust visual-semantic embeddings.
\newblock In {\em 2017 IEEE International Conference on Computer Vision
  (ICCV)}, pages 3591--3600. IEEE, 2017.

\bibitem{izsl}
W.~Wang, Y.~Pu, V.~K. Verma, K.~Fan, Y.~Zhang, C.~Chen, P.~Rai, and L.~Carin.
\newblock Zero-shot learning via class-conditioned deep generative models.
\newblock In {\em Thirty-Second AAAI Conference on Artificial Intelligence},
  2018.

\bibitem{cub200}
P.~Welinder, S.~Branson, T.~Mita, C.~Wah, F.~Schroff, S.~Belongie, and
  P.~Perona.
\newblock {Caltech-UCSD Birds 200}.
\newblock Technical Report CNS-TR-2010-001, California Institute of Technology,
  2010.

\bibitem{cbam}
S.~Woo, J.~Park, J.-Y. Lee, and I.~So~Kweon.
\newblock Cbam: Convolutional block attention module.
\newblock In {\em Proceedings of the European Conference on Computer Vision
  (ECCV)}, pages 3--19, 2018.

\bibitem{fgn}
Y.~Xian, T.~Lorenz, B.~Schiele, and Z.~Akata.
\newblock Feature generating networks for zero-shot learning.
\newblock In {\em Proceedings of the IEEE conference on computer vision and
  pattern recognition}, pages 5542--5551, 2018.

\bibitem{agzsl}
H.~Zhang, Y.~Long, L.~Liu, and L.~Shao.
\newblock Adversarial unseen visual feature synthesis for zero-shot learning.
\newblock {\em Neurocomputing}, 329:12--20, 2019.

\bibitem{holes}
Z.~Zhang, Y.~Song, and H.~Qi.
\newblock Age progression/regression by conditional adversarial autoencoder.
\newblock In {\em Proceedings of the IEEE Conference on Computer Vision and
  Pattern Recognition}, pages 5810--5818, 2017.

\bibitem{rkt}
B.~Zhao, X.~Sun, X.~Hong, Y.~Yao, and Y.~Wang.
\newblock Zero-shot learning via recurrent knowledge transfer.
\newblock In {\em 2019 IEEE Winter Conference on Applications of Computer
  Vision (WACV)}, pages 1308--1317. IEEE, 2019.

\end{thebibliography}
}

\end{document}